\documentclass{article}
\usepackage{ijcai26}

\usepackage{times}
\usepackage{soul}
\usepackage{url}
\usepackage[hidelinks]{hyperref}
\usepackage[utf8]{inputenc}
\usepackage[small]{caption}
\usepackage{graphicx}
\usepackage{amsmath}
\usepackage{amsthm}
\usepackage{amssymb}
\usepackage{booktabs}
\usepackage{algorithm}
\usepackage{algorithmic}
\usepackage{multirow}

\usepackage{tikz}
\usetikzlibrary{arrows.meta,shapes.geometric}

\makeatletter
\renewcommand{\section}{\@startsection{section}{1}{\z@}%
  {-0.6ex \@plus -0.2ex \@minus -0.1ex}%
  {0.4ex \@plus 0.2ex}%
  {\normalfont\Large\bfseries}}

\renewcommand{\subsection}{\@startsection{subsection}{2}{\z@}%
  {-0.2ex \@plus -0.1ex \@minus -0.1ex}%
  {0.3ex \@plus 0.1ex}%
  {\normalfont\large\bfseries}}

\renewcommand{\subsubsection}{\@startsection{subsubsection}{3}{\z@}%
  {-0.1ex \@plus -0.05ex \@minus -0.05ex}%
  {0.2ex \@plus 0.1ex}%
  {\normalfont\normalsize\bfseries}}
\makeatother

\title{Knowledge-Integrated Representation Learning for Crypto Anomaly Detection under Extreme Label Scarcity; Relational Domain-Logic Integration with Retrieval-Grounded Context and Path-Level Explanations}

\title{Knowledge-Integrated Representation Learning for Crypto Anomaly Detection under Extreme Label Scarcity; Relational Domain-Logic Integration with Retrieval-Grounded Context and Path-Level Explanations}

\author{
Gyuyeon Na$^{1}$\thanks{These authors contributed equally to this work.}
\and
Minjung Park$^{2}$\footnotemark[1]
\and
Soyoun Kim$^{1}$\footnotemark[1]
\and
Jungbin Shin$^{1}$
\and
Sangmi Chai$^{1,3}$\thanks{Corresponding author.}\\
\affiliations
$^{1}$AI and Business Analytics, Ewha Womans University, Seoul, Republic of Korea\\
$^{2}$Department of Business Administration, Kumoh National Institute of Technology, Gumi, Republic of Korea\\
$^{3}$Coretrustlink, Seoul, Republic of Korea\\
\emails
amy-na@ewha.ac.kr,
mjpark@kumoh.ac.kr,
sykim07@ewha.ac.kr,
patra33@ewha.ac.kr ,
smchai@ewha.ac.kr
}

\begin{document}

\maketitle

\begin{abstract}
Detecting anomalous trajectories in decentralized crypto-networks is fundamentally challenged by extreme label scarcity and the adaptive evasion strategies of illicit actors. 
While Graph Neural Networks (GNNs) effectively capture local structural patterns, they struggle to internalize multi-hop, logic-driven motifs—such as fund dispersal and layering—that characterize sophisticated money laundering, limiting their forensic accountability under regulations like the FATF Travel Rule.

To address this limitation, we propose \emph{Relational Domain-Logic Integration (RDLI)}, a framework that embeds expert-derived heuristics as differentiable, logic-aware latent signals within representation learning. 
Unlike static rule-based approaches, RDLI enables the detection of complex transactional flows that evade standard message passing. 
To further account for market volatility, we incorporate a \emph{Retrieval-Grounded Context (RGC)} module that conditions anomaly scoring on regulatory and macroeconomic context, mitigating false positives caused by benign regime shifts. Under extreme label scarcity (0.01\%), RDLI outperforms state-of-the-art GNN baselines by 28.9\% in F1-score. 
A micro-expert user study ($n=24$) further confirms that RDLI’s path-level explanations significantly improve trustworthiness, perceived usefulness, and clarity compared to existing methods ($p<0.001$), highlighting the importance of integrating domain logic with contextual grounding for both accuracy and explainability.

\end{abstract}

\section{Introduction}

The rapid proliferation of decentralized finance (DeFi) has introduced unprecedented challenges to global financial stability, necessitating a paradigm shift in anomaly detection systems. Under the updated FATF Travel Rule, cryptocurrency service providers are no longer judged solely on detection accuracy; they are now mandated to provide interpretable and verifiable rationales for every flagged transaction. Consequently, modern anomaly detection has evolved from a purely predictive task into a multi-dimensional challenge that demands both high precision and regulatory-grade auditability.

However, satisfying these dual requirements is particularly arduous in the cryptocurrency domain due to two structural barriers. First, \textbf{\emph{extreme label scarcity}} persists as a defining characteristic: illicit actors continuously evolve their tactics, rendering historical labels rapidly obsolete and leaving only a minute fraction of reliable ground truth for training. Second, although contemporary Graph Neural Networks (GNNs) are effective at capturing localized interactions, they frequently suffer from inherent \textbf{\emph{``black-box'' limitations}}. As a result, they fail to decode long-horizon, logic-driven motifs—such as complex fund dispersal and layering patterns—that are emblematic of sophisticated financial malfeasance. This disconnect creates a critical mismatch between purely data-driven modeling approaches and the practical interpretive needs of forensic investigators.

To bridge this gap, we propose \textbf{\emph{Relational Domain-Logic Integration (RDLI)}}, a novel framework that transcends the limitations of standard message-passing architectures by embedding expert-derived heuristics directly into the representation learning process. Our key contributions are three-fold:

\begin{itemize}
    \item \textbf{Logic-Aware Latent Signals.} We formalize expert typologies into differentiable, logic-aware latent signals, enabling the recovery of multi-hop flow dynamics that conventional GNNs frequently bypass.
    \item \textbf{Retrieval-Grounded Context (RGC).} We introduce a contextual conditioning module that integrates real-time regulatory updates and macroeconomic shifts, effectively filtering out \emph{benign regime changes} that commonly induce false positives in volatile cryptocurrency markets.
    \item \textbf{Path-Level Explainability.} Unlike post-hoc attribution methods such as SHAP, RDLI generates causal, path-level explanations that directly map flagged activities to salient subgraphs and domain-logic cues, ensuring audit-ready transparency.
\end{itemize}
Rigorous evaluations on large-scale datasets demonstrate that, even under extreme label scarcity (0.01\%), RDLI achieves an F1-score improvement of 28.9\% over GNN-based baselines. In addition, a simulated forensic review with a micro-expert panel ($n=24$) shows that the proposed approach significantly improves \emph{Trustworthiness (TR)} and \emph{Perceived Usefulness (PU)} compared to conventional feature-centric explanations, aligning with emerging regulatory and domain requirements~\cite{takei2024fatf,chen2025health}. 

Together, these results underscore the effectiveness of fusing domain-specific logic with contextual grounding to support robust, explainable, and institution-ready financial monitoring systems.

\section{Related Work}

\subsection{Anomaly Detection under Structural Scarcity}

Anomaly detection in financial transaction systems is fundamentally constrained by severe label scarcity, as anomalous activities are rare and often confirmed only through delayed investigations~\cite{Chandola2009,Carcillo2021Scarff}. This challenge is further compounded by the rapid evolution of illicit strategies, which renders historical labels obsolete and undermines the assumptions of supervised learning in operational settings~\cite{Guidotti2019,Weber2019AMLBitcoin}.

Recent work shows that graph-based detection models suffer from representation collapse and unstable generalization when label availability falls below critical thresholds~\cite{Dou2020,wang2021gnn_fraud_survey}. Even with data augmentation or semi-supervised extensions, performance often saturates under extreme scarcity~\cite{chalapathy2022deep,Hu2023Transaction}. 

While some approaches leverage structural flow patterns or heuristic-driven signals to mitigate these limitations~\cite{Li2020FlowScope,cheng2025graph}, they typically assume sufficient data completeness and fail to maintain robustness when both labels and reliable priors are scarce. These observations motivate the need for alternative paradigms that can sustain stable generalization under conditions of structural scarcity.

\subsection{Knowledge-Integrated \& Explainable Detection}

To overcome data limitations, recent studies have explored the incorporation of expert heuristics into learning-based anomaly detection models. While some approaches treat expert rules as weak supervision or manually engineered features~\cite{xu2024challenges}, they often incur high maintenance costs and exhibit limited adaptability to evolving adversarial strategies~\cite{korycki2023adversarial}. In parallel, retrieval-augmented learning (RAL) methods have been proposed to enrich model predictions with external context; however, they primarily improve instance-level awareness and fail to structurally encode expert knowledge required for path-level reasoning~\cite{lewis2020retrieval}.

From a regulatory compliance perspective, anomaly detection systems must also provide auditable and verifiable rationales. Prevailing post-hoc explanation techniques such as SHAP offer only coarse-grained feature attribution and are inherently incapable of capturing the sequential intent underlying money laundering behaviors~\cite{Lundberg2017SHAP}. Although recent logic-based approaches have begun to address these challenges, they largely remain confined to closed-world assumptions and static knowledge representations, limiting their applicability in dynamic, real-world financial environments. These limitations highlight the need for frameworks that jointly integrate expert heuristics with external unstructured context to support both robust detection and regulatory-aligned explainability.

\section{Proposed Method}

\begin{figure*}[ht]
    \centering
    \includegraphics[width=\textwidth, height=0.35\textheight, keepaspectratio]{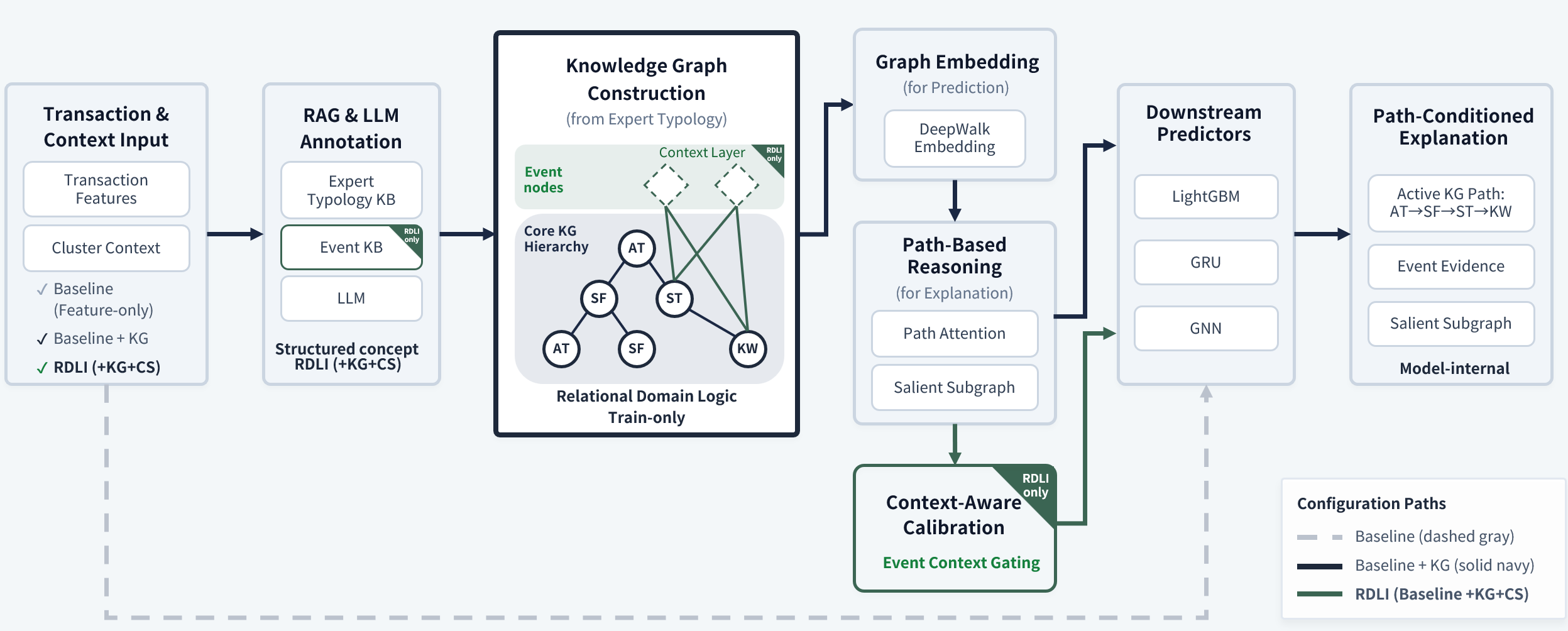}
    \caption{Overview of the RDLI framework}
    \label{fig:rdli_framework}
\end{figure*}

We propose a unified \emph{knowledge-guided anomaly detection framework}, termed \textbf{\emph{Relational Domain-Logic Integration (RDLI)}}. As shown in Figure~\ref{fig:rdli_framework}, RDLI encodes expert knowledge into a shared representation layer that is utilized by multiple downstream predictors.

\subsection{Expert Knowledge Graph Construction}

We employ Large Language Models (LLMs) to automatically generate structured annotations from raw transaction contexts based on expert-defined typologies. These annotations are subsequently transformed into an \emph{Expert Knowledge Graph} designed to capture hierarchical domain logic with high semantic density.

We formalize the Expert Knowledge Graph as a heterogeneous directed graph
$\mathcal{G} = (\mathcal{V}, \mathcal{E}, \mathcal{T}, \phi)$,
where $\mathcal{V}$ denotes the set of nodes and $\mathcal{E}$ denotes the set of directed edges encoding hierarchical dependencies.
The node type mapping function $\phi: \mathcal{V} \rightarrow \mathcal{T}$ assigns each node to a semantic category
$\mathcal{T} = \{\mathrm{AT}, \mathrm{SF}, \mathrm{ST}, \mathrm{KW}\}$,
corresponding to \emph{Anomaly Type}, \emph{Subtype Family}, \emph{Subtype}, and \emph{Keyword}, respectively.

Edges are constructed to enforce a fixed hierarchical structure
$\mathrm{AT} \rightarrow \mathrm{SF} \rightarrow \mathrm{ST} \rightarrow \mathrm{KW}$,
ensuring that high-level anomaly concepts are progressively specialized into fine-grained evidential descriptors.

For a transaction $x_i$, we define an induced logic path $P_i$ as an ordered hierarchical sequence:
\[
P_i = \bigl[
v_{\mathrm{AT}}^{(i)},\;
v_{\mathrm{SF}}^{(i)},\;
v_{\mathrm{ST}}^{(i)},\;
\{v_{\mathrm{KW}}^{(i,k)}\}_{k=1}^{K_i}
\bigr],
\]
where $v \in \mathcal{V}$ denotes the concept nodes associated with transaction $x_i$, and $K_i$ represents the number of keyword-level evidential nodes activated for that transaction.

\subsubsection{LLM-based Structured Annotation Generation}
\label{subsec:llm_annotation}

In this study, expert typologies are incorporated into a LLM via carefully designed prompts, enabling the automatic generation of \emph{structured, transaction-level annotations} from raw transaction and cluster contexts. We employ \emph{Gemini-2.5-Flash} to produce controlled hierarchical semantic labels, keywords, and concise explanations, which are stored alongside original transaction attributes in a unified schema. The annotation process required 3,378 seconds and consumed 1,585,415 tokens, after which the generated texts were embedded using \emph{text-embedding-004} for retrieval and similarity analysis.

\paragraph{Train-only Construction and Leakage Prevention}
To enforce strict forecasting principles, the concept graph is constructed exclusively from the training split. Test transactions are projected into the pre-trained \emph{DeepWalk} embedding space, with unseen concepts mapped to zero vectors to prevent information leakage. In addition, label-proximal fields (e.g., \emph{Anomaly Type}) are excluded from direct predictive inputs and used only for structural context modeling.

\subsection{Event-based Context Modeling}

We propose a ~\textit{Retrieval-Grounded Context (RGC)} module to filter benign regime changes. The system extracts narratives from news APIs (e.g., GNews) to form a searchable corpus. Each transaction is vectorized using the Gemini embedding model, and relevant context is retrieved via cosine similarity: This retrieves external evidence (e.g., regulatory warnings) to enrich the latent representation.

\[
\text{sim}(q, d) = \frac{q \cdot d}{\|q\| \cdot \|d\|}
\]

\label{sec:model_arch}

\begin{algorithm}[tb]
\caption{RAG-Augmented LLM Annotation with Concept Graph Embedding and Prediction}
\label{alg:llm_concept}

\footnotesize
\textbf{Input}: Transaction features $X=\{x_i\}$, cluster context $C=\{c_i\}$, expert KB $\mathcal{K}$, event KB $\mathcal{E}$ \\
\textbf{Parameters}: Train index set $\mathcal{I}_{tr}$, test index set $\mathcal{I}_{te}$, top-$k$ retrieval $(k_K, k_E)$ \\
\textbf{Output}: Anomaly score or label $\hat{y}_i$ for each transaction

\begin{algorithmic}[1]
\STATE Initialize empty annotation set $\mathcal{A} \leftarrow \emptyset$

\FOR{each transaction $i$}
    \STATE $q_i \leftarrow \mathrm{BuildQuery}(x_i, c_i)$
    \STATE $\mathcal{K}_i \leftarrow \mathrm{RetrieveTopK}(q_i, \mathcal{K}, k_K)$ \COMMENT{expert insights (typology chunks)}
    \STATE $\mathcal{E}_i \leftarrow \mathrm{RetrieveTopK}(q_i, \mathcal{E}, k_E)$ \COMMENT{event evidence (event chunks)}
    \STATE $a_i \leftarrow \mathrm{LLMAnnotate}(x_i, c_i, \mathcal{K}_i, \mathcal{E}_i)$
    \STATE Extract $(\mathrm{AT}_i, \mathrm{SF}_i, \mathrm{ST}_i, \{\mathrm{KW}_{i,1..k}\})$ from $a_i$
    \STATE $\mathcal{A} \leftarrow \mathcal{A} \cup \{a_i\}$
\ENDFOR

\STATE Initialize train-only concept graph $G_{tr} \leftarrow (\mathcal{V}_{tr}, \mathcal{E}_{tr})$
\FOR{each $i \in \mathcal{I}_{tr}$}
    \STATE Add nodes $\{\mathrm{AT}_i, \mathrm{SF}_i, \mathrm{ST}_i, \mathrm{KW}_{i,j}\}$ to $\mathcal{V}_{tr}$
    \STATE Add edges $\mathrm{AT}_i \rightarrow \mathrm{SF}_i \rightarrow \mathrm{ST}_i \rightarrow \mathrm{KW}_{i,j}$ to $\mathcal{E}_{tr}$
\ENDFOR

\STATE $Z \leftarrow \mathrm{DeepWalk}(G_{tr})$ \COMMENT{concept-level embeddings}

\FOR{each transaction $i$}
    \STATE $e_i \leftarrow \mathrm{Pool}(Z, a_i)$
    \STATE $r_i \leftarrow \mathrm{Concat}(x_i, a_i^{cat}, e_i)$
\ENDFOR

\STATE Train predictor $f_\theta$ on $\{(r_i, y_i)\}_{i \in \mathcal{I}_{tr}}$
\FOR{each $i \in \mathcal{I}_{te}$}
    \STATE $\hat{y}_i \leftarrow f_\theta(r_i)$
\ENDFOR

\STATE \textbf{return} $\{\hat{y}_i\}$
\end{algorithmic}
\end{algorithm}

\subsection{Model Architecture}
This study adopts the unified knowledge-guided anomaly detection framework outlined in Algorithm~\ref{alg:llm_concept}.

Let $\mathbf{h}_v \in \mathbb{R}^d$ be the embedding vector of a node $v$ learned via DeepWalk. The semantic representation $\mathbf{e}_i$ for transaction $i$ is computed via an attention-weighted pooling over its associated logic path $P_i$:$$\mathbf{e}_i = \sum_{v \in P_i} \alpha_v \cdot \mathbf{h}_v$$where $\alpha_v$ denotes the relevance weight of concept $v$.Finally, the unified input vector $\mathbf{r}_i$ for downstream predictors is constructed by concatenating the raw feature vector $\mathbf{x}_i$, the logic embedding $\mathbf{e}_i$, and the retrieved context embedding $\mathbf{c}_i$:$$\mathbf{r}_i = \text{Concat}(\mathbf{x}_i, \mathbf{e}_i, \mathbf{c}_i)$$

\subsubsection{Unified Knowledge-Guided Input Representation}
\label{subsec:unified_input}
For each transaction, we construct a unified feature representation with two components:
(i) \emph{numeric and categorical transaction attributes} (e.g., value, direction, self-transfer flags), with categorical variables one-hot encoded; and
(ii) an optional \emph{semantic embedding} derived from the expert knowledge graph by pooling \emph{DeepWalk embeddings} of the transaction’s structured concepts.

Depending on the experimental setting, the semantic component is either \emph{empty} (\emph{feature-only}), a \emph{graph-based embedding} (\emph{KG}), or a KG embedding augmented via \emph{retrieval-based alignment} (\emph{KG+RAG}). All models share the same unified input format.

\subsubsection{Tree-based Predictor }
\label{subsec:lgbm}

We employ LightGBM as a strong tree-based baseline for tabular anomaly detection, following prior studies that adopt gradient-boosted decision trees for transaction-level anomaly modeling~\cite{jia2025lmae4eth}. 
The model takes the unified input representation as a fixed-length feature vector and learns non-linear decision boundaries through gradient-boosted decision trees. 
As LightGBM does not explicitly model temporal or relational dependencies, it serves as a reference point to evaluate the benefits of sequential and graph-aware architectures.

\subsubsection{Neural Sequence Predictor }
\label{subsec:na_model}

To capture temporal dependencies across transactions, we adopt a gated recurrent unit (GRU)-based neural anomaly model, following prior work on structural--temporal modeling for proactive anomaly detection~\cite{Park2025HyPVLEAD}. 
Transaction streams are segmented into fixed-length sliding windows, each represented as a sequence of unified input vectors. 
The GRU encoder processes each window and summarizes it into a latent representation via the final hidden state, which is subsequently passed to a sigmoid prediction head to estimate anomaly probability ~\cite{li2019reading,tan2021cooperative}.

This architecture enables the model to detect anomalous behavior that emerges over multiple consecutive transactions, rather than from isolated events.

\subsubsection{Graph Neural Network Predictor}
\label{subsec:gnn}

To model relational dependencies among transactions and addresses, we adopt a graph neural network (GNN)-based predictor following prior spatio-temporal graph learning frameworks~\cite{Ghaffari2025STMGraph}. Transactions or addresses are represented as nodes, while edges encode interaction or fund-flow relationships among them. We utilize GraphSAGE~\cite{hamilton2017inductive} as the GNN backbone to support inductive learning on evolving transaction graphs. Through fixed-size neighborhood sampling and aggregation, GraphSAGE enables efficient generalization to nodes that emerge in the temporal evaluation split.

Each node is initialized with a unified, knowledge-guided feature representation, allowing the GNN to jointly exploit structural dependencies and expert-informed semantic signals. 
Compared to sequence-based models, GNNs explicitly model relational and spatial structures that are difficult to capture using time-series representations alone, making them well-suited for anomaly detection in evolving graph settings~\cite{wang2019robust,pareja2020evolvegcn,zhao2021csgnn}.

\subsection{Explanation Scoring}

To generate an explanation, we identify the salient subgraph by maximizing the alignment score between the transaction context and the knowledge path. The explanation score $S(P_i)$ is formulated as:$$S(P_i) = \max_{P \in \mathcal{N}(i)} \left( \lambda_1 \cdot \text{sim}(\mathbf{x}_i, P) + \lambda_2 \cdot \text{sim}(\mathcal{E}_i, P) \right)$$where $\mathcal{N}(i)$ is the set of candidate paths and $\mathcal{E}_i$ is the retrieved event context.

\section{Experimental Setup}
Detailed dataset construction specifications and hyperparameter settings are provided in the supplementary material. To ensure reproducibility, the source code and the stratified dataset subset are included in the supplementary submission.
\subsection{Dataset and Prerocessing}

\begin{table}[t]
\centering
\small
\begin{tabular}{p{2.0cm} p{5.8cm}}
\hline
Dataset / Component & Details \\
\hline
LLM-Based Transaction Annotation
&
\textit{Subset}: Stratified 0.01\% of transactions \par
\textit{LLM Role}: Semantic inference (non-labeling) \par
\textit{Key Fields}: anomaly\_type, subtype\_family, subtype \par
direction, is\_self\_transfer, coin\_infer \par
coin, year, source\_file \par
abs\_usd\_value, \_dt, date \par
annotation, keyword1--keyword5 \par
llm\_raw \par
Expert-guided semantic enrichment used for KG construction. \\
\hline
Event Dataset
&
\textit{Purpose}: Contextual grounding of anomalies \par
\textit{Key Fields}: event\_id, event\_title, event\_date \par
published\_at, anomaly\_type, coin \par
description, gate\_reason, text\_sample \par
is\_anomaly, llm\_verified \par
Aligns detected transaction patterns with real-world incidents. \\
\hline
Expert Knowledge Dataset
&
\textit{Format}: Structured CSV knowledge base \par
\textit{Key Fields}: chunk\_id, chunk\_text \par
sheet\_name, row\_idx \par
Integrated via RAG to retrieve expert-defined typologies and explanations \par
Ensures domain-grounded and interpretable annotations. \\
\hline
\end{tabular}
\caption{Overview of LLM-annotated transaction data, event context, and expert knowledge datasets}
\label{tab:dataset_overview}
\end{table}

Experiments were conducted on a real-world cryptocurrency transaction dataset spanning from January 2020 to December 2024. To simulate extreme label scarcity while preserving the severe class imbalance observed in operational settings, we extracted a stratified subset containing only 0.01\% of labeled instances. The dataset encompasses a diverse range of anomalous behaviors, including mixing activities (e.g., CoinJoin and Tornado Cash) and scam-related events (e.g., rug pulls). Detailed dataset statistics are summarized in Table~\ref{tab:dataset_overview}.

To prevent information leakage and ensure a realistic evaluation setting, we applied both address-disjoint and temporal train--test splitting strategies. This experimental design follows established best practices in cryptocurrency anomaly detection and logic-aware learning under non-stationary environments~\cite{hoffman2018metrics,jia2025lmae4eth,li2025association,lei2025llmcrypto,sun2025ethereum,shyalika2025nsf,thimonier2024retrieval}.

\subsubsection{Raw Transaction Dataset}
This raw dataset serves as the foundational input for subsequent annotation,
knowledge enrichment, and model training.

\subsubsection{LLM-Based Annotation and Knowledge Graph Enrichment}

To incorporate domain expertise and enhance semantic interpretability, we enrich a stratified 0.01\% subset of transactions using an LLM guided by expert-defined instructions and anomaly typologies. The resulting annotations are used to construct a transaction-centric knowledge graph that enables structured reasoning and explainable anomaly detection.

This design is grounded in classical research on knowledge representation and logical abstraction~\cite{levesque:functional-foundations,levesque:belief,abelson-et-al:scheme,brachman1989overview,gottlob:nonmon,nebel:jair-2000,baumgartner2001visual,gls:hypertrees}.

\subsubsection{Event Dataset}

An external event dataset was incorporated to provide contextual grounding. The event dataset enables alignment between observed transaction patterns
and known real-world incidents, supporting context-aware interpretation
of detected anomalies.

\subsubsection{Expert Knowledge Dataset}

Expert knowledge was compiled into a structured CSV-based knowledge base and integrated into a \emph{Retrieval-Augmented Generation (RAG)} framework. This allows the LLM to retrieve relevant expert-defined typologies and descriptions during the annotation process.

To ensure leakage prevention, fair model comparison, and alignment with \emph{realistic operational settings}, we applied the following sampling and splitting strategies:

\begin{enumerate}
    \item \textbf{Address-disjoint sampling:}
    Wallet addresses were strictly separated between training and evaluation sets to prevent information leakage and reflect deployment conditions.

    \item \textbf{Temporal consistency:}
    Transactions were chronologically ordered, with earlier data assigned to training, ensuring no future information was used.

    \item \textbf{Stratified sampling:}
    The original class imbalance between normal and anomalous transactions was preserved.

    \item \textbf{Anomaly type coverage:}
    Sampling constraints ensured minimal representation of major anomaly categories, avoiding bias toward specific types.

    \item \textbf{Fixed sampling for fair comparison:}
    A single \emph{0.01\%} sampled subset was extracted once and reused across all experiments to ensure consistent comparison.
\end{enumerate}

\subsection{Configurations and Evaluation}

We compare three configurations: Feature-only (Baseline), Feature + KG, and RDLI (Feature + KG + CS). Performance is evaluated using F1-score and Recall, prioritizing the minimization of missed detections in compliance settings.

\begin{enumerate}
    \item \textbf{Feature-only (Baseline)}:
    Uses raw transaction features with binary anomaly labels
    (0: normal, 1: anomalous), serving as a reference without external knowledge.

    \item \textbf{Feature + KG}:
    Augments baseline features with expert knowledge graph embeddings.
    The knowledge graph encodes domain heuristics and typology-based semantic
    relationships, which are transformed into vector representations using
    DeepWalk.

    \item \textbf{RDLI (Feature + KG + CS)}:
    The full proposed model that integrates knowledge graph embeddings with
    contextual signals.

\end{enumerate}

To evaluate performance under extreme label scarcity, we conduct experiments on both the full dataset and a stratified 0.01\% subset. This design minimizes reliance on data scale while explicitly validating the effectiveness of knowledge integration. Class imbalance and anomaly-type coverage are preserved via stratified sampling.

Given the severe imbalance inherent in anomaly detection, we report macro-averaged \emph{Accuracy}, \emph{Precision}, \emph{Recall}, \emph{F1-score}, and \emph{AUC-ROC}, with \emph{Recall} prioritized to minimize missed anomalies. This setting reflects real-world cryptocurrency monitoring, where transaction dynamics and adversarial behaviors evolve continuously.

To assess robustness and generalizability, we further evaluate the model on the Kaggle \emph{Credit Card Transactions Dataset}~\cite{card_kaggle_dataset} under identical conditions (0.01\% labels, temporal split). The results demonstrate that ~\textit{RDLI} generalizes beyond cryptocurrency networks and maintains stable performance across heterogeneous financial domains.

\subsection{Experimental Design for Simulated Explanation Evaluation}

We conduct a simulated user study to examine how different explanation formats for the same anomalous cryptocurrency transaction influence user perception. A \emph{micro-expert panel} ($n=24$) was recruited, consisting of professionals with experience in virtual asset transactions, anomaly detection, and regulatory or compliance analysis.

Participants reviewed two realistic anomaly scenarios involving complex transaction flows and address relationships. For each scenario, they were presented with two explanation formats for the same detection outcome: (i) a \emph{baseline feature-based explanation} summarizing observable transaction attributes, and (ii) an \emph{RDLI explanation} that integrates structured expert knowledge and contextual information into a natural-language rationale.

Explanation quality was evaluated across seven constructs—\emph{Trust (TR)}, \emph{Perceived Usefulness (PU)}, \emph{Perceived Ease of Use (PEOU)}, \emph{Consistency (CON)}, \emph{Explainability (EXP)}, \emph{Clarity (CLA)}, and \emph{Acceptance Intention (AI)}—using three to five items per construct on a five-point Likert scale. Survey items were adapted from validated instruments in prior work on explainable systems and usability, and comprehension checks were included to ensure response validity.

\section{Experimental Results}
\subsection{Performance under Extreme Label Scarcity}

This experiment examines the impact of \emph{extreme label scarcity} on existing anomaly detection models and evaluates whether the proposed approach can mitigate this limitation. When trained on the full dataset, baseline models (LightGBM, GRU, and GNN) achieve stable performance, indicating that data-driven learning is effective under sufficient supervision. However, under severe label reduction (0.01\%), all baseline models suffer substantial performance degradation, particularly in Recall and F1-score. This highlights the vulnerability of conventional anomaly detection methods in realistic low-label regimes, where rare anomalous behaviors are difficult to identify with limited supervision.

In contrast, the proposed model integrates \emph{\textbf{expert-heuristic knowledge}} via a \emph{\textbf{knowledge graph (KG)}} and \emph{\textbf{contextual signals (CS)}}, maintaining robust performance even under extreme label scarcity. By explicitly encoding domain expertise and transaction-level context, it enables anomaly detection beyond purely data-driven patterns.

As shown in Table~\ref{tab:performance_lowlabel}, the proposed approach consistently improves \emph{Recall} and \emph{F1-score} over corresponding baselines trained under identical low-label conditions. These results suggest that anomaly detection performance is not solely driven by label volume, but critically depends on the systematic integration of expert heuristics and contextual information, particularly in regulated cryptocurrency environments with scarce and evolving labels.

\begin{table}[t]
\centering
\renewcommand{\arraystretch}{1.15}
\resizebox{\columnwidth}{!}{
\normalsize
\begin{tabular}{llccccc}
\hline
Backbone & Model Variant & Acc. & Prec. & Rec. & F1 & AUC \\
\hline
\multirow{4}{*}{LightGBM}
& Feature-only (Full)
& 0.7870 & 0.8408 & 0.7870 & 0.8018 & 0.8518 \\
& Feature-only (0.01\%)
& 0.7892 & 0.7796 & 0.7892 & 0.7825 & 0.8110 \\
& RDLI (KG + CS)
& 0.9963 & 0.9950 & 1.0000 & 0.9975 & 0.9996 \\
& $\Delta$ (RDLI -- Full)
& 0.2093 & 0.1542 & 0.2130 & 0.1957 & 0.1478 \\
\hline

\multirow{4}{*}{GRU}
& Feature-only (Full)
& 0.7271 & 0.3999 & 0.5395 & 0.4593 & 0.6768 \\
& Feature-only (0.01\%)
& 0.4139 & 0.8516 & 0.2437 & 0.3789 & 0.6883 \\
& RDLI (KG + CS)
& 0.7389 & 0.8898 & 0.7351 & 0.8051 & 0.8290 \\
& $\Delta$ (RDLI -- Full)
& 0.0118 & 0.4899 & 0.1956 & 0.3458 & 0.1522 \\
\hline

\multirow{4}{*}{GNN}
& Feature-only (Full)
& 0.9043 & 0.7460 & 0.5291 & 0.6920 & 0.8438 \\
& Feature-only (0.01\%)
& 0.6815 & 0.6472 & 0.6815 & 0.6243 & 0.6793 \\
& RDLI (KG + CS)
& 0.9129 & 0.9342 & 0.8235 & 0.8923 & 0.8823 \\
& $\Delta$ (RDLI -- Full)
& 0.0086 & 0.1882 & 0.2944 & 0.2003 & 0.0385 \\
\hline

\end{tabular}
}
\caption{Performance comparison of RDLI under extreme label scarcity (0.01\%).
}
\label{tab:performance_lowlabel}
\end{table}

\subsection{Ablation Study}

To identify the sources of performance gains, we conduct an ablation study across three backbones—\emph{LightGBM}, \emph{GRU}, and \emph{GNN}—by incrementally adding or removing components under identical training conditions.

Across all backbones, \emph{feature-only} configurations exhibit clear limitations under extreme label scarcity. This suggests that performance degradation arises not from insufficient model capacity, but from the \textbf{absence of informative decision signals} under limited supervision.

Incorporating \emph{structural information} yields partial improvements, particularly for GRU and GNN models, indicating that relational patterns provide useful anomaly cues. However, these gains remain inconsistent across metrics, implying that \textbf{structure alone is insufficient} to resolve decision ambiguity.

The full configuration consistently achieves the most stable and balanced performance across all backbones, with \emph{GNN-based models} benefiting most strongly. Overall, the results demonstrate that performance gains stem from the \textbf{complementary integration of multiple components} rather than any single module. Under extreme label scarcity, effective anomaly detection therefore depends more on principled integration of \emph{structural} and \emph{contextual} information than on labeled data volume.

\begin{table}[t]
\centering
\renewcommand{\arraystretch}{1.15}
\resizebox{\columnwidth}{!}{%
\begin{tabular}{llccccc}
\hline
Backbone & Model Variant & Acc. & Prec. & Rec. & F1 & AUC \\
\hline
\multirow{3}{*}{LightGBM}
& Feature-only & 0.7892 & 0.7796 & 0.7892 & 0.7825 & 0.8110 \\
& KG          & 0.9808 & 0.9939 & 0.9799 & 0.9799 & 0.9994 \\
& RDLI (KG + CS) & 0.9963 & 0.9950 & 1.0000 & 0.9975 & 0.9996 \\
\hline
\multirow{3}{*}{GRU}
& Feature-only & 0.4139 & 0.8516 & 0.2437 & 0.3789 & 0.6883 \\
& KG          & 0.6306 & 0.8304 & 0.6239 & 0.7125 & 0.6783 \\
& RDLI (KG + CS) & 0.7389 & 0.8898 & 0.7351 & 0.8051 & 0.8290 \\
\hline
\multirow{3}{*}{GNN}
& Feature-only & 0.6815 & 0.6472 & 0.6815 & 0.6243 & 0.6793 \\
& KG          & 0.7086 & 0.6889 & 0.7124 & 0.6622 & 0.7318 \\
& RDLI (KG + CS) & 0.9129 & 0.9342 & 0.8235 & 0.8923 & 0.8823 \\
\hline
\end{tabular}%
}
\caption{Ablation study under extreme label scarcity (0.01\%) on the cryptocurrency transaction dataset}
\label{tab:ablation_unified}
\end{table}

\subsection{Reproducibility Analysis on Card Transaction Data}

To assess \textit{reproducibility and cross-domain robustness}, we extended our evaluation to the publicly available Kaggle \emph{Credit Card Transactions Dataset}~\cite{card_kaggle_dataset}, under the same \emph{extreme label scarcity} setting (0.01\%). This experiment examines how RDLI behaves when the data topology shifts from decentralized crypto-flows to centralized banking transactions.

\paragraph{Performance on Tabular Baselines}
The LightGBM-based RDLI configuration exhibits strong generalization, achieving an F1-score improvement of 7.65\% over the feature-only baseline. This result indicates that even in traditional financial domains, injecting \emph{logic-aware} embeddings effectively sharpens decision boundaries in tabular models.

\paragraph{Performance Divergence in Deep Models}
For sequence-based (GRU) and graph-based (GNN) backbones, RDLI consistently improves performance, though absolute gains remain lower than in the cryptocurrency domain. We attribute this divergence to a \emph{topological mismatch} between the two domains:

\begin{itemize}
    \item \textbf{Episodic vs.\ Sequential.} Credit card fraud is typically episodic, lacking the long-horizon dependencies required for GRUs to fully leverage sequential modeling.
    \item \textbf{Star-Graph vs.\ Flow-Network.} Card transactions form shallow bipartite star graphs (User--Merchant), constraining GNN message passing compared to the deep flow networks of blockchains.
\end{itemize}

\paragraph{Compliance Implications}
Despite these constraints, RDLI achieves a \textit{perfect Recall of 1.0} across all backbones. Given that missed fraud incurs significantly higher regulatory risk than false positives, this property highlights the framework’s \emph{risk-averse suitability} for compliance-oriented deployments.

\paragraph{Discussion}
Overall, this study delineates RDLI’s operating envelope: while it consistently enhances tabular baselines across domains, its full potential emerges in settings with \emph{complex relational logic} and \emph{deep sequential dependencies}, such as DeFi and AML investigations, where purely data-driven models often fail.

\begin{table}[t]
\centering
\renewcommand{\arraystretch}{1.15}
\normalsize
\resizebox{\columnwidth}{!}{%
\begin{tabular}{llccccc}
\hline
Backbone & Model Variant & Acc. & Prec. & Rec. & F1 & AUC \\
\hline
\multirow{3}{*}{LightGBM}
& Feature-only & 0.9230 & 0.8461 & 0.9166 & 0.8800 & 0.9413 \\
& KG          & 0.9487 & 0.9166 & 0.9166 & 0.9166 & 0.9259 \\
& RDLI (KG + CS) & 0.9744 & 1.0000 & 0.9166 & 0.9565 & 0.9320 \\
\hline
\multirow{3}{*}{GRU}
& Feature-only & 0.9994 & 0.3333 & 1.0000 & 0.5000 & 0.9994 \\
& KG          & 0.9974 & 0.1111 & 1.0000 & 0.2000 & 0.9974 \\
& RDLI (KG + CS) & 0.9929 & 0.0430 & 1.0000 & 0.0833 & 0.9929 \\
\hline
\multirow{3}{*}{GNN}
& Feature-only & 0.9822 & 0.2534 & 0.9384 & 0.3970 & 0.9355 \\
& KG          & 0.9945 & 0.0833 & 1.0000 & 0.1538 & 0.9946 \\
& RDLI (KG + CS) & 0.9972 & 0.2842 & 1.0000 & 0.4424 & 0.9984 \\
\hline
\end{tabular}%
}

\caption{Ablation study under extreme label scarcity (0.01\%) on the card transaction dataset.}
\label{tab:ablation_card}
\end{table}
\subsection{Explainability Analysis}

\begin{figure}[ht]
    \centering
    \includegraphics[width=\columnwidth]{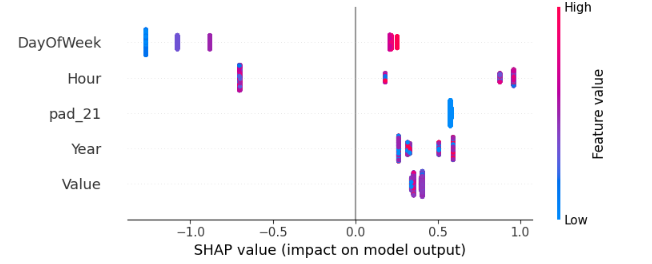}
    \caption{SHAP-based top-5 feature importance for the LightGBM (Full) model.}
    \label{fig:shap}
\end{figure}

To benchmark the interpretability of standard anomaly detection models, we employ SHAP (SHapley Additive exPlanations), a widely adopted method for explaining predictions generated by ``black-box'' models~\cite{Lundberg2017SHAP}. 
Figure~\ref{fig:shap} visualizes the top-5 feature attributions for the LightGBM (Full) model.

\paragraph{The Illusion of Interpretability}
At first glance, SHAP appears effective in highlighting global feature importance, identifying attributes such as \emph{Value}, \emph{Year}, and \emph{DayOfWeek} as dominant predictors. However, within a forensic auditing context, this feature-centric representation reveals a critical semantic gap.

\begin{itemize}
    \item \textbf{Atomistic vs.\ Relational Reasoning.} SHAP decomposes a model prediction into independent feature contributions. In contrast, money laundering behavior is inherently relational: a high-value transaction is rarely suspicious in isolation, but rather due to its position within a broader transactional structure, such as a dispersal or ``fan-out'' pattern. SHAP is fundamentally unable to encode this sequential and relational logic.
    
    \item \textbf{Ambiguity of Intent.} As illustrated in Figure~\ref{fig:shap}, distinct fraud typologies often exhibit overlapping SHAP value distributions. For instance, a high-value signal (red dot) may correspond equally to a legitimate VIP transfer, a theft-related movement, or a layering operation. Without explicit access to the underlying behavioral motif, auditors cannot reliably distinguish legitimate high-volume trading from illicit laundering activities.
\end{itemize}

\paragraph{The Need for Path-Level Reasoning}
Consequently, while feature attribution methods such as SHAP provide statistical justification for model outputs, they fail to deliver the behavioral rationale demanded by regulatory frameworks such as the FATF Travel Rule. This limitation directly motivates the design of RDLI, which shifts the explanatory focus from isolating ``suspicious features'' to reconstructing the complete ``suspicious path.'' We empirically validate this hypothesis through the subsequent micro-expert user study.

\subsection{Validating Operational Utility with Domain Experts}

To assess the practical viability of the proposed framework, we conducted a simulated forensic review with a micro-expert panel ($n=24$) consisting of professionals in anti-money laundering (AML). Participants were asked to review explanations for flagged anomalies under two conditions: a \textit{Baseline (Feature-centric)} explanation and an \textit{RDLI (Path-centric)} explanation.

As reported in Table~5, RDLI significantly outperformed the baseline across all evaluated psychometric constructs, with paired $t$-tests indicating statistical significance at the $p < 0.001$ level. Qualitative feedback further revealed that baseline explanations imposed a high cognitive burden, requiring experts to manually infer behavioral intent from fragmented feature lists. In contrast, RDLI’s narrative, path-level structure functioned as a \emph{cognitive scaffold}, closely aligning with investigators’ mental models and forensic reasoning processes. These results confirm that RDLI delivers the level of \emph{audit-readiness} required for effective human--AI collaborative workflows in regulated financial environments.

\begin{table}[ht]
\centering
\footnotesize
\setlength{\tabcolsep}{4pt}  
\begin{tabular}{c c c c}
\toprule
Anomaly Case & Measure & t-value & p-value \\
\midrule
1 & TR    & -2.346 & $< .004$ \\
1 & PU    & -3.342 & $< .001$ \\
1 & PEOU  & -3.974 & $< .001$ \\
1 & CON   & -5.003 & $< .001$ \\
1 & EXP   & -6.859 & $< .001$ \\
1 & CLA   & -2.421 & $< .012$ \\
\midrule
2 & TR    & -2.298 & $< .001$ \\
2 & PU    & -4.243 & $< .001$ \\
2 & PEOU  & -3.015 & $< .021$ \\
2 & CON   & -4.463 & $< .001$ \\
2 & EXP   & -5.972 & $< .001$ \\
2 & CLA   & -3.194 & $< .003$ \\
\bottomrule
\end{tabular}
\caption{Paired $t$-test results comparing baseline \& RDLI explanations}

\label{tab:rdli_ttest}
\end{table}

\section{Conclusion}

This study addresses a central challenge in cryptocurrency anomaly detection: reconciling the demand for \emph{\textbf{audit-ready transparency}} under regulations such as the FATF Travel Rule with the opacity and data sparsity of modern deep learning models. We argue that in high-stakes financial environments, \textbf{purely data-driven approaches are insufficient} to capture the adaptive nature of financial crime.

To bridge this gap, we proposed \emph{\textbf{Relational Domain-Logic Integration (RDLI)}}, a framework that embeds \textbf{expert heuristics as differentiable latent signals} and anchors them with retrieval-grounded market context. Rather than memorizing surface-level patterns, \emph{RDLI} reconstructs the underlying \emph{\textbf{logic of crime}} through structured reasoning.

Empirically, \emph{RDLI} demonstrates \textbf{strong resilience under extreme label scarcity (0.01\%)}, outperforming state-of-the-art GNN baselines by 28.9\% in F1-score. Additional experiments on credit card transaction data and a micro-expert evaluation further confirm its \textbf{robustness across financial domains} and alignment with human expert reasoning.

Overall, this work advocates a shift from correlation-driven black-box models toward neuro-symbolic financial AI that explicitly integrates domain logic and context. We position \emph{RDLI} as a step toward \textbf{explainable, trustworthy, and regulation-compliant} autonomous financial monitoring systems.

\bibliographystyle{named}
\bibliography{ijcai26}

\end{document}